\documentclass{llncs}
\usepackage{graphicx}
\usepackage{float}
\usepackage{subfigure}
\usepackage{booktabs}
\usepackage{hyperref}
\usepackage{rotating} 
\usepackage{caption}
\usepackage{multirow}
\usepackage{longtable}
\usepackage{array}

\begin{document}

\title{Overview and Results: CL-SciSumm Shared Task 2019}
\titlerunning{CL-SciSumm 2019}  
%
\author{Muthu Kumar Chandrasekaran \inst{1} \and Michihiro Yasunaga \inst{2} 
\and \\ Dragomir Radev \inst{2} \and Dayne Freitag \inst{1} \and Min-Yen Kan\inst{3}}
\authorrunning{Chandrasekaran et al.} 
\institute{SRI International, USA
\and
Yale University, USA
\and
School of Computing, National University of Singapore, Singapore 
\\
\email{\tt{cmkumar087@gmail.com}}
}

\maketitle              

\begin{abstract}
The CL-SciSumm Shared Task is the first medium-scale shared task on
scientific document summarization in the computational
linguistics~(CL) domain. In 2019, it comprised three tasks: (1A)
identifying relationships between citing documents and the referred
document, (1B) classifying the discourse facets, and (2) generating
the abstractive summary. The dataset comprised 40 annotated sets of
citing and reference papers of the 
CL-SciSumm 2018 corpus and 1000 more 
from the SciSummNet dataset.
All papers are from the open access research papers in
the CL domain. This overview describes the participation and the
official results of the CL-SciSumm 2019 Shared Task, organized as a
part of the $42^{nd}$ Annual Conference of the Special Interest Group
in Information Retrieval (SIGIR), held in Paris, France in July 2019.
We compare the participating systems in terms of two evaluation
metrics and discuss the use of ROUGE as an evaluation metric. The
annotated dataset used for this shared task and the scripts used for
evaluation can be accessed and used by the community at:
\url{https://github.com/WING-NUS/scisumm-corpus}.
\end{abstract}


\section{Introduction}
\label{s:intro}
CL-SciSumm explores summarization of scientific research in the domain of computational 
linguistics research. It encourages the incorporation of new kinds of information in 
automatic scientific paper summarization, such as the facets of research information 
being summarized in the research paper. CL-SciSumm also encourages the use of citing 
mini-summaries written in other papers, by other scholars, when they refer to the paper. 
The Shared Task dataset comprises the set of citation sentences (i.e., ``citances'') that 
reference a specific paper as a (community-created) summary of a topic or 
paper~\cite{qazvinian2008scientific}. Citances for a reference paper are considered a 
synopses of its key points and also its key contributions and importance within an 
academic community~\cite{nakov2004citances}. The advantage of using citances is that they 
are embedded with meta-commentary and offer a contextual, interpretative layer to the 
cited text. Citances offer a view of the cited paper which could complement the reader's 
context, possibly as a scholar~\cite{sparckjones2007automatic}.

The CL-SciSumm Shared Task is aimed at bringing together the summarization community to 
address challenges in scientific communication summarization. Over time, we anticipate 
that the Shared Task will spur the creation of new resources, tools and evaluation 
frameworks. 

A pilot CL-SciSumm task was conducted at TAC 2014, as part of the larger BioMedSumm Task\footnote{http://www.nist.gov/tac/2014}. 
In 2016, a second CL-Scisumm Shared Task \cite{jaidka2017insights} was held as part of the Joint Workshop on Bibliometric-enhanced Information Retrieval and Natural Language Processing for Digital Libraries (BIRNDL) workshop~\cite{mayr16:_editorial} at the Joint Conference on Digital Libraries (JCDL 2016). This paper provides the results and insights from CL-SciSumm 2017, which was held as part of subsequent BIRNDL 2017 workshop\cite{MayrCJ17} at the annual ACM Conference on Research and Development in Information Retrieval (SIGIR\footnote{\url{http://sigir.org/sigir2017/}}).


\section{Task}

CL-SciSumm defined two serially dependent tasks that participants could attempt, 
given a canonical training and testing set of papers. \\

\label{s:task}
\noindent \textbf{Given}: A topic consists of a Reference Paper (RP) and ten or more
Citing Papers (CPs) that all contain citations to the RP. In each CP, the 
text spans (i.e., citances) have been identified that pertain to a particular 
citation to the RP.
Additionally, the dataset provides three types of summaries for each RP:
\begin{itemize}  
	\item {the abstract, written by the authors of the research paper.}
    \item {the community summary, collated from the reference spans of its 
    		citances.}
  	\item {a human-written summary, written by the annotators of the CL-SciSumm 
    		annotation effort.}
\end{itemize}

\noindent \textbf{Task 1A}: For each citance, identify the spans of text (cited text 
spans) in the RP that most accurately reflect the citance. These are of the 
granularity of a sentence fragment, a full sentence, or several consecutive 
sentences (no more than 5). \\

\noindent \textbf{Task 1B}: For each cited text span, identify what facet of the paper 
it belongs to, from a predefined set of facets. \\

\noindent \textbf{Task 2}: Finally, generate a structured summary of the RP from the 
cited text spans of the RP. The length of the summary should not exceed 250 
words. This was an optional bonus task.

\section{Development}
\label{s:dev}
We built the CL-SciSumm corpus by randomly sampling research papers (Reference papers, 
RPs) from the ACL Anthology corpus and then downloading the citing papers (CPs) for those 
which had at least ten citations. The prepared dataset then comprised annotated citing 
sentences for a research paper, mapped to the sentences in the RP which they referenced. 
Summaries of the RP were also included. 

The CL-SciSumm 2019 corpus consisted for 40 annotated RPs 
and their CPs. These are the same as described in our 
overview paper in CL-SciSumm 2018~\cite{jaidka2018overview}.
The test set was blind. We reused the blind test 
we used for CL-SciSumm 2018 since we want to have a comparable 
evaluation CL-SciSumm 2019 systems that will have 
additional training data (see Section~\ref{ss:annot}). 

For details of the general procedure followed to construct the CL-SciSumm 
corpus, and changes made to the procedure in CL-SciSumm-2016, please see
\cite{jaidka2017insights}. In 2017, we made revisions to the 
corpus to remove citances from passing citations. These are described in 
\cite{jaidka2017overview}.


\subsection{Annotation}
\label{ss:annot}

The first annotated CL-SciSumm corpus was 
released for The CL-SciSumm 16 shared task. This was 
annotated based on annotation scheme from what was 
followed in previous editions of the task and the original 
BiomedSumm task developed by Cohen et. al\footnote{\url{http://www.nist.gov/tac/2014}}: 
Given each RP and its associated CPs, the annotation group was 
instructed to find citations to the RP in each CP. Specifically, 
the citation text, citation marker, reference text, and discourse 
facet were identified for each citation of the RP found in the CP.

Then CL-Scisumm-17 and CL-Scisumm-18 incrementally added 
more annotated RPs to its current size of 40 annotated RPs. 

For CL-Scisumm-19, we augment this dataset 
both Task 1a and Task 2 so that they have approximately 1000 
data points as opposed to 40 in previous years. 
Specifically, for Task 1, we used the method proposed by~\cite{nomoto2018resolving} 
to prepare noisy training data for about 1000 unannotated papers. 
This method involves automatically matching a citance in a CP with 
approximately similar reference spans in its RPs. The number of 
reference spans per citance is a hyperparameter that can set as 
input. For Task 2, we used the SciSummNet corpus proposed by~\cite{yasunaga&al.aaai19.scisumm}.

\section{Overview of Approaches}
\label{s:methods}

Nine systems out of the seventeen registered systems --
in Task~1 and a subset of five also participated in Task~2 -- 
submitted their output for 
evaluation. We include these system papers in the BIRNDL 2019 proceedings. 
We will now briefly summarise their methods and key results in lexicographic 
order by team name. \\

\textbf{System~1} is from Nanjing University of Science and Technology~\cite{system_1}. 
For Task 1A, they use multi-classifiers and integrate their results via 
voting system. Compared with previous work, this year they make new selection
of features based on correlation analysis, apply similarity-based negative sampling 
strategy when creating training dataset and add deep learning models for classifications.
For Task 1B, they firstly calculate the probability that each word would belong 
to the specific facet based on training corpus and then some prior rules are 
added to obtain final result. For Task 2, to obtain a logical summary, they group 
sentences in two ways, first based on their relevance between abstract segments 
and second arranged by recognized facet from task 1B. Then they pick out 
important sentences via ranking.

\textbf{System~2} is from Beijing University of Posts and Telecommunications (BUPT) \cite{system_2}. 
They build a new feature of Word2vec\_H for the CNN model to calculate sentence similarity for citation linkage. In addition to the methods used last year, they also intend to apply CNN for facet classification. In order to improve the performance of summarization, they develop more semantic representations for sentences based on neural network language models to construct new kernel matrix used in Determinantal Point Processes (DPPs).

\textbf{System~3} is from University of Manchester~\cite{system_3}. 
For Task~1 they looked into supervised and semi-supervised approaches. They explored 
the potential of fine-tuning bidirectional transformers for the identification of cited 
passages. They further formalised the task as a similarity ranking problem and 
implemented bilateral multi-perspective matching for natural language sentences. 
For Task 2, they used hybrid summarisation methods to create a summary from the content 
of the paper and the cited text spans.

\textbf{System~4} is from University of Toulouse~ \cite{system_4}.
They focus on Task 1A. They first identify candidate sentences in the reference paper and compute 
their similarities to the citing sentence using tf-idf and embedding-based methods as well as other 
features such as POS tags. They submitted 15 runs with different configurations.

\textbf{System~7} is from IIIT Hyderabad and Adobe Research~\cite{system_7}. 
Their architecture incorporates transfer learning by utilising a combination 
of pretrained embeddings which are subsequently used for building models for 
the given tasks. In particular, for task 1A, they locate the related text spans 
referred to by the citation text by creating paired text representations and 
employ pre-trained embedding mechanisms in conjunction with XGBoost, a gradient 
boosted decision tree algorithm to identify textual entailment. For task 1B, 
they make use of the same pretrained embeddings and use the RAKEL algorithm 
for multi-label classification. 

\textbf{System~8} is from Universitat Pompeu Fabra and Universidad de la 
Republica~\cite{system_8}. They propose a supervised system based on recurrent 
neural networks and an unsupervised system based on sentence similarity for Task 1A, 
one supervised approach for Task 1B, and one supervised approach for Task 2. 
The approach for Task 2 follows the method by the winning approach in CL-SciSumm 2018.

\textbf{System~9} is from Politecnico di Torino~\cite{system_9}.
Their approach to tasks 1A and 1B relies on an ensemble of classification and 
regression models trained on the annotated pairs of cited and citing sentences. 
Facet assignment is based on the relative positions of the cited sentences locally to the corresponding section and globally in the entire paper. Task 2 is addressed by 
predicting the overlap (in terms of units of text) between the selected text 
spans and the summary generated by the domain experts. The output summary 
consists of the subset of sentences maximizing the predicted overlap score.

\textbf{System~12} is from Nanjing University and Kim Il Sung University~\cite{system_12}.
They propose a novel listwise ranking method for cited text identification. Their method have two stages: similarity-based ranking and supervised listwise ranking. In the first stage, we select the top-5 sentences per a citation text, due to the modified Jaccard similarity. These top-5 selected sentences are proceeded to rank by a CitedListNet (listwise ranking model based on deep learning). They select 36 similarity features and 11 section information as feature. Finally, they select two sentences on the sentence list ranked by CitedList- Net.

\textbf{System~17} is from National Technical University of Athens, Athens University of Economics 
and Business, and Athena Research and Innovation Center~\cite{system_17}.
Their approach is twofold. Firstly they classify sentences of an abstract to predefined classes called ``zones''. They use sentences from selected zones to find the most similar ones of the rest sentences of the paper which constitute the ``candidate sentences''. Secondly, they employ a siamese bi-directional GRU neural network with a logistic regression layer to classify if a citation sentence cites a candidate sentence.

\section{Evaluation}

An automatic evaluation script was used to measure system performance for 
\textbf{Task~1A}, in terms of the sentence ID overlaps between the sentences 
identified in system output, versus the gold standard created by human annotators. 
The raw number of overlapping sentences were used to calculate the precision, 
recall and $F_1$ score for each system. We followed the approach in most 
SemEval tasks in reporting the overall system performance as its micro-averaged 
performance over all topics in the blind test set.


Additionally, we calculated lexical overlaps in terms of the ROUGE-2 and 
ROUGE-SU4 scores~\cite{lin2004rouge} between the system 
output and the human annotated gold standard reference spans.

We have been reporting ROUGE scoring since CL-SciSumm 17, for 
Tasks~1a and Task~2.
Recall-Oriented Understudy for Gisting Evaluation
(ROUGE) is a set of metrics used to automatically evaluate
summarization systems~\cite{lin2004rouge} by measuring the overlap
between computer-generated summaries and multiple human written
reference summaries. In previous studies, ROUGE scores have
significantly correlated with human judgments on summary
quality~\cite{liu2008correlation}. Different variants of ROUGE differ
according to the granularity at which overlap is calculated.  For
instance, ROUGE--2 measures the bigram overlap between the candidate
computer-generated summary and the reference summaries. More
generally, ROUGE--N measures the $n$-gram overlap. ROUGE--L measures
the overlap in Longest Common Subsequence (LCS). ROUGE--S measures
overlaps in skip-bigrams or bigrams with arbitrary gaps
in-between. ROUGE-SU uses skip-bigram plus unigram overlaps.
CL-SciSumm 2017 uses ROUGE-2 and ROUGE-SU4 for its evaluation.\\

\textbf{Task~1B} was evaluated as a proportion of the correctly classified
discourse facets by the system, contingent on the expected response of
Task~1A. As it is a multi-label classification, this task was also
scored based on the precision, recall and $F_1$ scores.

\textbf{Task~2} was optional, and also evaluated using the ROUGE--2 and
ROUGE--SU4 scores between the system output and three types of gold
standard summaries of the research paper: the reference paper's
abstract, a community summary, and a human summary.

The evaluation scripts have been provided at the CL-SciSumm Github
repository\footnote{\url{github.com/WING-NUS/scisumm-corpus}} where
the participants may run their own evaluation and report the results.

\section{Results}
\label{s:sysruns}

\begin{figure}
  \centering
  \subfigure[]{\includegraphics[width=\textwidth]{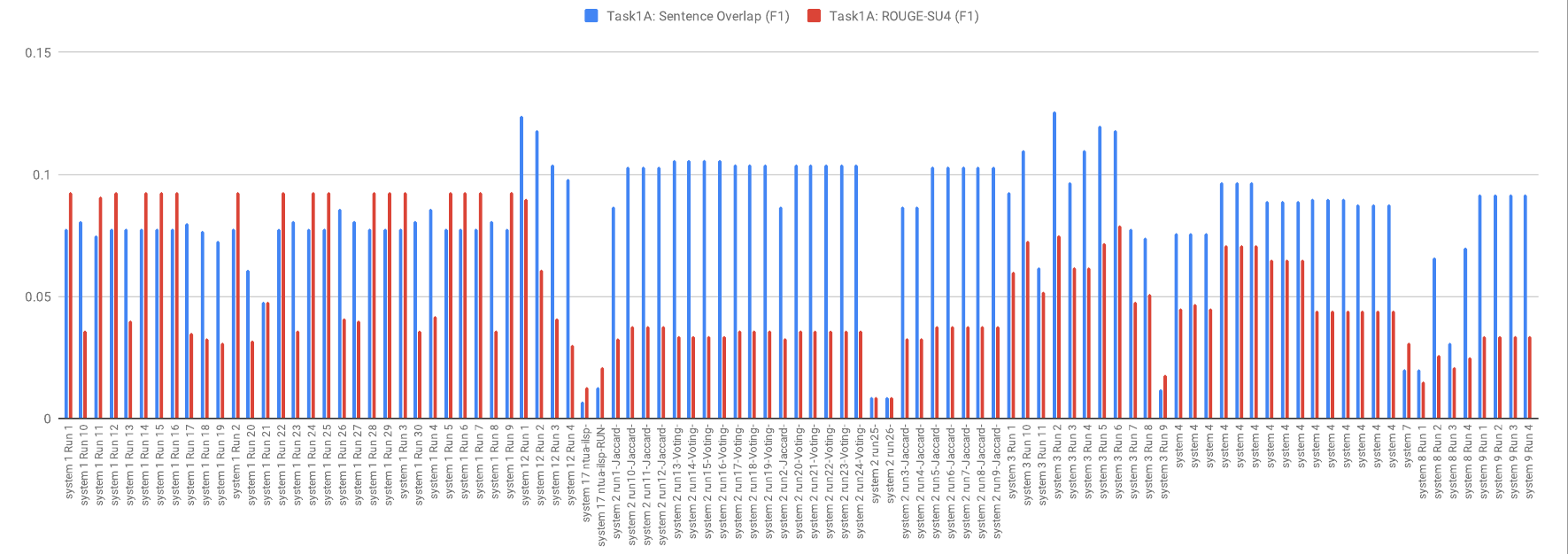}\label{fig:exph}}
  \subfigure[]{\includegraphics[width=\textwidth]{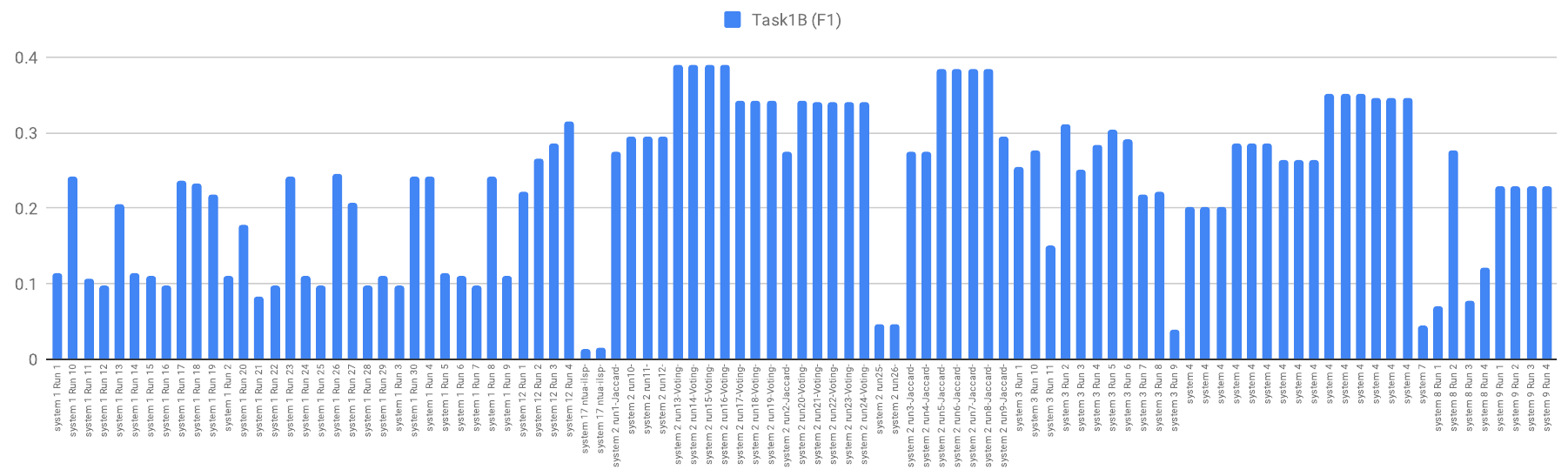}\label{fig:exgr}}
  \caption{Performances on (a) Task~1A in terms of sentence overlap
    and ROUGE-SU4, and (b) Task~1B conditional on Task~1A}
  \label{fig:task1}
\end{figure}

This section compares the participating systems in terms 
of their performance. Five of the nine system that did 
Task~1 also did the bonus Task~2. Following are the plots 
with their performance measured by ROUGE--2 and 
ROUGE--SU4 against the 3 gold standard summary 
types. The results are provided in Table~\ref{tab:task1} 
and Figure~\ref{fig:task1}. The detailed implementation of 
the individual runs are described in the system papers 
included in this proceedings volume.
\begin{table}
\renewcommand{\arraystretch}{1.15}
\centering
\scalebox{0.5}{
\begin{tabular}
{|p{11cm}| >{\raggedright\arraybackslash}p{3.5cm}
| >{\raggedright\arraybackslash}p{3.5cm}
| >{\raggedright\arraybackslash}p{3.5cm}
|
}
\hline
\textbf{System} & \shortstack{\textbf{Task 1A: Sentence}\\\textbf{Overlap ($F_1$)}} & \shortstack{\textbf{Task 1A:}\\\textbf{ROUGE-SU4 $F_1$}} & \textbf{Task 1B}\\
\hline
system 3 Run 2                                                          & 0.126                         & 0.075                  & 0.312       \\
system 12 Run 1                                                         & 0.124                         & 0.090                  & 0.221       \\
system 3 Run 5                                                          & 0.120                         & 0.072                  & 0.303       \\
system 3 Run 6                                                          & 0.118                         & 0.079                  & 0.292       \\
system 12 Run 2                                                         & 0.118                         & 0.061                  & 0.266       \\
system 3 Run 10                                                         & 0.110                         & 0.073                  & 0.276       \\
system 3 Run 4                                                          & 0.110                         & 0.062                  & 0.283       \\
system 2 run15-Voting-1.1-SubtitleAndHfw-QD\_method\_1                  & 0.106                         & 0.034                  & 0.389       \\
system 2 run13-Voting-1.1-SubtitleAndHfw-LSA\_method\_3                 & 0.106                         & 0.034                  & 0.389       \\
system 2 run14-Voting-1.1-SubtitleAndHfw-LSA\_method\_4                 & 0.106                         & 0.034                  & 0.389       \\
system 2 run16-Voting-1.1-SubtitleAndHfw-SentenceVec\_method\_2         & 0.106                         & 0.034                  & 0.389       \\
system 2 run23-Voting-2.0-Voting-QD\_method\_1                          & 0.104                         & 0.036                  & 0.341       \\
system 2 run24-Voting-2.0-Voting-SentenceVec\_method\_2                 & 0.104                         & 0.036                  & 0.341       \\
system 2 run20-Voting-2.0-TextCNN-SentenceVec\_method\_2                & 0.104                         & 0.036                  & 0.342       \\
system 2 run21-Voting-2.0-Voting-LSA\_method\_3                         & 0.104                         & 0.036                  & 0.341       \\
system 2 run18-Voting-2.0-TextCNN-LSA\_method\_4                        & 0.104                         & 0.036                  & 0.342       \\
system 2 run22-Voting-2.0-Voting-LSA\_method\_4                         & 0.104                         & 0.036                  & 0.341       \\
system 2 run19-Voting-2.0-TextCNN-QD\_method\_1                         & 0.104                         & 0.036                  & 0.342       \\
system 2 run17-Voting-2.0-TextCNN-LSA\_method\_3                        & 0.104                         & 0.036                  & 0.342       \\
system 12 Run 3                                                         & 0.104                         & 0.041                  & 0.286       \\
system 2 run10-Jaccard-Focused-Voting-LSA\_method\_4                    & 0.103                         & 0.038                  & 0.294       \\
system 2 run7-Jaccard-Focused-SubtitleAndHfw-QD\_method\_1              & 0.103                         & 0.038                  & 0.385       \\
system 2 run5-Jaccard-Focused-SubtitleAndHfw-LSA\_method\_3             & 0.103                         & 0.038                  & 0.385       \\
system 2 run9-Jaccard-Focused-Voting-LSA\_method\_3                     & 0.103                         & 0.038                  & 0.294       \\
system 2 run12-Jaccard-Focused-Voting-SentenceVec\_method\_2            & 0.103                         & 0.038                  & 0.294       \\
system 2 run6-Jaccard-Focused-SubtitleAndHfw-LSA\_method\_4             & 0.103                         & 0.038                  & 0.385       \\
system 2 run11-Jaccard-Focused-Voting-QD\_method\_1                     & 0.103                         & 0.038                  & 0.294       \\
system 2 run8-Jaccard-Focused-SubtitleAndHfw-SentenceVec\_method\_2     & 0.103                         & 0.038                  & 0.385       \\
system 12 Run 4                                                         & 0.098                         & 0.030                  & 0.315       \\
system 3 Run 3                                                          & 0.097                         & 0.062                  & 0.251       \\
system 4 WithoutEmb\_Training20182019\_Test2019\_3\_0.1                 & 0.097                         & 0.071                  & 0.286       \\
system 4 WithoutEmb\_Training2018\_Test2019\_3\_0.1                     & 0.097                         & 0.071                  & 0.286       \\
system 4 WithoutEmb\_Training2019\_Test2019\_3\_0.1                     & 0.097                         & 0.071                  & 0.286       \\
system 3 Run 1                                                          & 0.093                         & 0.060                  & 0.255       \\
system 9 Run 2                                                          & 0.092                         & 0.034                  & 0.229       \\
system 9 Run 3                                                          & 0.092                         & 0.034                  & 0.229       \\
system 9 Run 1                                                          & 0.092                         & 0.034                  & 0.229       \\
system 9 Run 4                                                          & 0.092                         & 0.034                  & 0.229       \\
system 4 WithoutEmbTopsim\_Training20182019\_Test2019\_0.15\_5\_0.05    & 0.090                         & 0.044                  & 0.351       \\
system 4 WithoutEmbTopsim\_Training2019\_Test2019\_0.15\_5\_0.05        & 0.090                         & 0.044                  & 0.351       \\
system 4 WithoutEmbTopsim\_Training2018\_Test2019\_0.15\_5\_0.05        & 0.090                         & 0.044                  & 0.351       \\
system 4 WithoutEmbPOS\_Training20182019\_Test2019\_3\_0.1              & 0.089                         & 0.065                  & 0.263       \\
system 4 WithoutEmbPOS\_Training2019\_Test2019\_3\_0.1                  & 0.089                         & 0.065                  & 0.263       \\
system 4 WithoutEmbPOS\_Training2018\_Test2019\_3\_0.1                  & 0.089                         & 0.065                  & 0.263       \\
system 4 WithoutEmbTopsimPOS\_Training2019\_Test2019\_0.15\_5\_0.05     & 0.088                         & 0.044                  & 0.346       \\
system 4 WithoutEmbTopsimPOS\_Training2018\_Test2019\_0.15\_5\_0.05     & 0.088                         & 0.044                  & 0.346       \\
system 4 WithoutEmbTopsimPOS\_Training20182019\_Test2019\_0.15\_5\_0.05 & 0.088                         & 0.044                  & 0.346       \\
system 2 run1-Jaccard-Cascade-Voting-LSA\_method\_3                     & 0.087                         & 0.033                  & 0.274       \\
system 2 run3-Jaccard-Cascade-Voting-QD\_method\_1                      & 0.087                         & 0.033                  & 0.274       \\
system 2 run4-Jaccard-Cascade-Voting-SentenceVec\_method\_2             & 0.087                         & 0.033                  & 0.274       \\
system 2 run2-Jaccard-Cascade-Voting-LSA\_method\_4                     & 0.087                         & 0.033                  & 0.274       \\
system 1 Run 26                                                         & 0.086                         & 0.041                  & 0.245       \\
system 1 Run 4                                                          & 0.086                         & 0.042                  & 0.241       \\
system 1 Run 30                                                         & 0.081                         & 0.036                  & 0.242       \\
system 1 Run 27                                                         & 0.081                         & 0.040                  & 0.207       \\
system 1 Run 8                                                          & 0.081                         & 0.036                  & 0.242       \\
system 1 Run 10                                                         & 0.081                         & 0.036                  & 0.242       \\
system 1 Run 23                                                         & 0.081                         & 0.036                  & 0.242       \\
system 1 Run 17                                                         & 0.080                         & 0.035                  & 0.236       \\
system 3 Run 7                                                          & 0.078                         & 0.048                  & 0.218       \\
system 1 Run 12                                                         & 0.078                         & 0.093                  & 0.098       \\
system 1 Run 15                                                         & 0.078                         & 0.093                  & 0.110       \\
system 1 Run 28                                                         & 0.078                         & 0.093                  & 0.098       \\
system 1 Run 2                                                          & 0.078                         & 0.093                  & 0.110       \\
system 1 Run 9                                                          & 0.078                         & 0.093                  & 0.110       \\
system 1 Run 25                                                         & 0.078                         & 0.093                  & 0.098       \\
system 1 Run 13                                                         & 0.078                         & 0.040                  & 0.205       \\
system 1 Run 24                                                         & 0.078                         & 0.093                  & 0.110       \\
system 1 Run 22                                                         & 0.078                         & 0.093                  & 0.098       \\
system 1 Run 3                                                          & 0.078                         & 0.093                  & 0.098       \\
system 1 Run 5                                                          & 0.078                         & 0.093                  & 0.113       \\
system 1 Run 6                                                          & 0.078                         & 0.093                  & 0.110       \\
system 1 Run 1                                                          & 0.078                         & 0.093                  & 0.113       \\
system 1 Run 14                                                         & 0.078                         & 0.093                  & 0.113       \\
system 1 Run 7                                                          & 0.078                         & 0.093                  & 0.098       \\
system 1 Run 16                                                         & 0.078                         & 0.093                  & 0.098       \\
system 1 Run 29                                                         & 0.078                         & 0.093                  & 0.110       \\
system 1 Run 18                                                         & 0.077                         & 0.033                  & 0.232       \\
system 4 unweightedPOS\_W2v\_Training2018\_Test2019\_3\_0.05            & 0.076                         & 0.045                  & 0.201       \\
system 4 unweightedPOS\_W2v\_Training20182019\_Test2019\_3\_0.05        & 0.076                         & 0.047                  & 0.201       \\
system 4 unweightedPOS\_W2v\_Training2019\_Test2019\_3\_0.05            & 0.076                         & 0.045                  & 0.201       \\
system 1 Run 11                                                         & 0.075                         & 0.091                  & 0.106       \\
system 3 Run 8                                                          & 0.074                         & 0.051                  & 0.221       \\
system 1 Run 19                                                         & 0.073                         & 0.031                  & 0.218       \\
system 8 Run 4                                                          & 0.070                         & 0.025                  & 0.122       \\
system 8 Run 2                                                          & 0.066                         & 0.026                  & 0.277       \\
system 3 Run 11                                                         & 0.062                         & 0.052                  & 0.150       \\
system 1 Run 20                                                         & 0.061                         & 0.032                  & 0.178       \\
system 1 Run 21                                                         & 0.048                         & 0.048                  & 0.083       \\
system 8 Run 3                                                          & 0.031                         & 0.021                  & 0.078       \\
system 8 Run 1                                                          & 0.020                         & 0.015                  & 0.070       \\
system 7                                                                & 0.020                         & 0.031                  & 0.045       \\
system 17 ntua-ilsp-RUN-NNT                                             & 0.013                         & 0.021                  & 0.016       \\
system 3 Run 9                                                          & 0.012                         & 0.018                  & 0.039       \\
system 2 run25-Word2vec-H-CNN-SubtitleAndHfw-QD\_method\_1              & 0.009                         & 0.009                  & 0.047       \\
system 2 run26-Word2vec-H-CNN-SubtitleAndHfw-SentenceVec\_method\_2     & 0.009                         & 0.009                  & 0.047       \\
system 17 ntua-ilsp-RUN\_NNF                                            & 0.007                         & 0.013                  & 0.013  \\\hline    
\end{tabular}}
\caption{
\scriptsize
Systems' performance in Task 1A and 1B, ordered by their 
$F_1$-scores for sentence overlap on Task 1A. Each system's rank by 
their performance on ROUGE on Task 1A and 1B are shown in 
parentheses.}
\label{tab:task1}
\end{table}


\begin{table}
\scalebox{0.48}{
\renewcommand{\arraystretch}{1.15}
\centering
\begin{tabular}{|p{12.5cm} | p{1.65cm} | p{1.65cm} | p{1.65cm} | p{1.65cm} | p{1.65cm} | p{1.65cm} |}
\hline
\multirow{2}{*}{\textbf{System}} & 
\multicolumn{2}{c}{} \textbf{Vs. Abstract} &
\multicolumn{2}{|c}{} \textbf{Vs. Community } &
\multicolumn{2}{|c}{} \textbf{Vs. Human} \\
\cline{2-7}
 & \textbf{R--2}& \textbf{RSU--4}& \textbf{R--2}& \textbf{RSU--4}& \textbf{R--2}& \textbf{RSU--4}\\
\hline
\hline
system 3 Run 1                                                                 & 0.514        & 0.295          & 0.106         & 0.062           & 0.265     & 0.180       \\
system 3 Run 11                                                                & 0.514        & 0.295          & 0.106         & 0.062           & 0.265     & 0.180       \\
system 3 Run 6                                                                 & 0.514        & 0.295          & 0.106         & 0.062           & 0.265     & 0.180       \\
system 3 Run 2                                                                 & 0.514        & 0.295          & 0.106         & 0.062           & 0.265     & 0.180       \\
system 3 Run 7                                                                 & 0.514        & 0.295          & 0.106         & 0.062           & 0.265     & 0.180       \\
system 3 Run 10                                                                & 0.514        & 0.295          & 0.106         & 0.062           & 0.265     & 0.180       \\
system 3 Run 8                                                                 & 0.514        & 0.295          & 0.106         & 0.062           & 0.265     & 0.180       \\
system 3 Run 5                                                                 & 0.514        & 0.295          & 0.106         & 0.062           & 0.265     & 0.180       \\
system 3 Run 3                                                                 & 0.514        & 0.295          & 0.106         & 0.062           & 0.265     & 0.180       \\
system 3 Run 4                                                                 & 0.514        & 0.295          & 0.106         & 0.062           & 0.265     & 0.180       \\
system 3 Run 9                                                                 & 0.514        & 0.295          & 0.106         & 0.062           & 0.265     & 0.180       \\
system 2 run3-Jaccard-Cascade-Voting-QD\_method\_1\_human                      & 0.389        & 0.210          & 0.122         & 0.063           & 0.278     & 0.200       \\
system 2 run3-Jaccard-Cascade-Voting-QD\_method\_1\_abstract                   & 0.389        & 0.210          & 0.122         & 0.063           & 0.278     & 0.200       \\
system 2 run23-Voting-2.0-Voting-QD\_method\_1\_human                          & 0.386        & 0.227          & 0.121         & 0.063           & 0.257     & 0.189       \\
system 2 run19-Voting-2.0-TextCNN-QD\_method\_1\_human                         & 0.386        & 0.227          & 0.121         & 0.063           & 0.257     & 0.189       \\
system 2 run19-Voting-2.0-TextCNN-QD\_method\_1\_abstract                      & 0.386        & 0.227          & 0.121         & 0.063           & 0.257     & 0.189       \\
system 2 run23-Voting-2.0-Voting-QD\_method\_1\_abstract                       & 0.386        & 0.227          & 0.121         & 0.063           & 0.257     & 0.189       \\
system 2 run15-Voting-1.1-SubtitleAndHfw-QD\_method\_1\_human                  & 0.381        & 0.211          & 0.119         & 0.062           & 0.267     & 0.191       \\
system 2 run15-Voting-1.1-SubtitleAndHfw-QD\_method\_1\_abstract               & 0.381        & 0.211          & 0.119         & 0.062           & 0.267     & 0.191       \\
system 2 run10-Jaccard-Focused-Voting-LSA\_method\_4\_community                & 0.368        & 0.186          & 0.096         & 0.053           & 0.252     & 0.170       \\
system 2 run2-Jaccard-Cascade-Voting-LSA\_method\_4\_community                 & 0.368        & 0.186          & 0.096         & 0.053           & 0.252     & 0.170       \\
system 2 run18-Voting-2.0-TextCNN-LSA\_method\_4\_community                    & 0.368        & 0.186          & 0.096         & 0.053           & 0.252     & 0.170       \\
system 2 run6-Jaccard-Focused-SubtitleAndHfw-LSA\_method\_4\_community         & 0.368        & 0.186          & 0.096         & 0.053           & 0.252     & 0.170       \\
system 2 run14-Voting-1.1-SubtitleAndHfw-LSA\_method\_4\_community             & 0.368        & 0.186          & 0.096         & 0.053           & 0.252     & 0.170       \\
system 2 run22-Voting-2.0-Voting-LSA\_method\_4\_community                     & 0.368        & 0.186          & 0.096         & 0.053           & 0.252     & 0.170       \\
system 2 run11-Jaccard-Focused-Voting-QD\_method\_1\_human                     & 0.367        & 0.201          & 0.121         & 0.062           & 0.258     & 0.184       \\
system 2 run7-Jaccard-Focused-SubtitleAndHfw-QD\_method\_1\_abstract           & 0.367        & 0.201          & 0.121         & 0.062           & 0.258     & 0.184       \\
system 2 run11-Jaccard-Focused-Voting-QD\_method\_1\_abstract                  & 0.367        & 0.201          & 0.121         & 0.062           & 0.258     & 0.184       \\
system 2 run7-Jaccard-Focused-SubtitleAndHfw-QD\_method\_1\_human              & 0.367        & 0.201          & 0.121         & 0.062           & 0.258     & 0.184       \\
system 9 Run 1                                                                 & 0.364        & 0.196          & 0.196         & 0.104           & 0.218     & 0.144       \\
system 9 Run 3                                                                 & 0.359        & 0.194          & 0.195         & 0.104           & 0.211     & 0.141       \\
system 9 Run 2                                                                 & 0.346        & 0.176          & 0.209         & 0.112           & 0.215     & 0.140       \\
system 2 run5-Jaccard-Focused-SubtitleAndHfw-LSA\_method\_3\_community         & 0.343        & 0.171          & 0.097         & 0.049           & 0.254     & 0.174       \\
system 2 run13-Voting-1.1-SubtitleAndHfw-LSA\_method\_3\_community             & 0.343        & 0.171          & 0.097         & 0.049           & 0.254     & 0.174       \\
system 2 run1-Jaccard-Cascade-Voting-LSA\_method\_3\_community                 & 0.343        & 0.171          & 0.097         & 0.049           & 0.254     & 0.174       \\
system 2 run9-Jaccard-Focused-Voting-LSA\_method\_3\_community                 & 0.343        & 0.171          & 0.097         & 0.049           & 0.254     & 0.174       \\
system 2 run21-Voting-2.0-Voting-LSA\_method\_3\_community                     & 0.343        & 0.171          & 0.097         & 0.049           & 0.254     & 0.174       \\
system 2 run17-Voting-2.0-TextCNN-LSA\_method\_3\_community                    & 0.343        & 0.171          & 0.097         & 0.049           & 0.254     & 0.174       \\
system 9 Run 4                                                                 & 0.340        & 0.174          & 0.206         & 0.111           & 0.208     & 0.138       \\
system 8 Run 1                                                                 & 0.329        & 0.172          & 0.149         & 0.090           & 0.241     & 0.171       \\
system 2 run12-Jaccard-Focused-Voting-SentenceVec\_method\_2\_abstract         & 0.318        & 0.171          & 0.142         & 0.075           & 0.239     & 0.167       \\
system 2 run12-Jaccard-Focused-Voting-SentenceVec\_method\_2\_human            & 0.318        & 0.171          & 0.142         & 0.075           & 0.239     & 0.167       \\
system 2 run8-Jaccard-Focused-SubtitleAndHfw-SentenceVec\_method\_2\_abstract  & 0.318        & 0.171          & 0.142         & 0.075           & 0.239     & 0.167       \\
system 2 run8-Jaccard-Focused-SubtitleAndHfw-SentenceVec\_method\_2\_human     & 0.318        & 0.171          & 0.142         & 0.075           & 0.239     & 0.167       \\
system 8 Run 2                                                                 & 0.316        & 0.167          & 0.169         & 0.101           & 0.245     & 0.169       \\
system 8 Run 3                                                                 & 0.311        & 0.156          & 0.153         & 0.093           & 0.252     & 0.170       \\
system 2 run20-Voting-2.0-TextCNN-SentenceVec\_method\_2\_abstract             & 0.296        & 0.152          & 0.128         & 0.067           & 0.252     & 0.177       \\
system 2 run24-Voting-2.0-Voting-SentenceVec\_method\_2\_human                 & 0.296        & 0.152          & 0.128         & 0.067           & 0.252     & 0.177       \\
system 2 run20-Voting-2.0-TextCNN-SentenceVec\_method\_2\_human                & 0.296        & 0.152          & 0.128         & 0.067           & 0.252     & 0.177       \\
system 2 run24-Voting-2.0-Voting-SentenceVec\_method\_2\_abstract              & 0.296        & 0.152          & 0.128         & 0.067           & 0.252     & 0.177       \\
system 1 Run 26                                                                & 0.296        & 0.145          & 0.193         & 0.108           & 0.224     & 0.150       \\
system 1 Run 4                                                                 & 0.294        & 0.144          & 0.191         & 0.108           & 0.235     & 0.151       \\
system 2 run4-Jaccard-Cascade-Voting-SentenceVec\_method\_2\_abstract          & 0.287        & 0.155          & 0.121         & 0.066           & 0.247     & 0.175       \\
system 2 run4-Jaccard-Cascade-Voting-SentenceVec\_method\_2\_human             & 0.287        & 0.155          & 0.121         & 0.066           & 0.247     & 0.175       \\
system 2 run16-Voting-1.1-SubtitleAndHfw-SentenceVec\_method\_2\_human         & 0.277        & 0.150          & 0.124         & 0.064           & 0.246     & 0.179       \\
system 2 run16-Voting-1.1-SubtitleAndHfw-SentenceVec\_method\_2\_abstract      & 0.277        & 0.150          & 0.124         & 0.064           & 0.246     & 0.179       \\
system 2 run25-Word2vec-H-CNN-SubtitleAndHfw-QD\_method\_1\_abstract           & 0.277        & 0.158          & 0.115         & 0.059           & 0.238     & 0.167       \\
system 2 run25-Word2vec-H-CNN-SubtitleAndHfw-QD\_method\_1\_human              & 0.277        & 0.158          & 0.115         & 0.059           & 0.238     & 0.167       \\
system 1 Run 8                                                                 & 0.277        & 0.137          & 0.200         & 0.115           & 0.229     & 0.151       \\
system 1 Run 30                                                                & 0.276        & 0.137          & 0.204         & 0.117           & 0.237     & 0.154       \\
system 1 Run 10                                                                & 0.276        & 0.137          & 0.204         & 0.117           & 0.237     & 0.154       \\
system 1 Run 18                                                                & 0.262        & 0.127          & 0.196         & 0.113           & 0.223     & 0.149       \\
system 2 run26-Word2vec-H-CNN-SubtitleAndHfw-SentenceVec\_method\_2\_human     & 0.261        & 0.145          & 0.126         & 0.066           & 0.222     & 0.153       \\
system 2 run26-Word2vec-H-CNN-SubtitleAndHfw-SentenceVec\_method\_2\_abstract  & 0.261        & 0.145          & 0.126         & 0.066           & 0.222     & 0.153       \\
system 8 Run 4                                                                 & 0.246        & 0.147          & 0.131         & 0.084           & 0.170     & 0.141       \\
system 1 Run 20                                                                & 0.239        & 0.122          & 0.177         & 0.102           & 0.231     & 0.158       \\
system 2 run15-Voting-1.1-SubtitleAndHfw-QD\_method\_1\_community              & 0.207        & 0.123          & 0.126         & 0.070           & 0.215     & 0.153       \\
system 2 run3-Jaccard-Cascade-Voting-QD\_method\_1\_community                  & 0.205        & 0.118          & 0.130         & 0.069           & 0.201     & 0.144       \\
system 2 run4-Jaccard-Cascade-Voting-SentenceVec\_method\_2\_community         & 0.204        & 0.123          & 0.140         & 0.077           & 0.221     & 0.159       \\
system 2 run24-Voting-2.0-Voting-SentenceVec\_method\_2\_community             & 0.203        & 0.126          & 0.138         & 0.076           & 0.225     & 0.164       \\
system 2 run20-Voting-2.0-TextCNN-SentenceVec\_method\_2\_community            & 0.203        & 0.126          & 0.138         & 0.076           & 0.225     & 0.164       \\
system 2 run8-Jaccard-Focused-SubtitleAndHfw-SentenceVec\_method\_2\_community & 0.199        & 0.115          & 0.131         & 0.073           & 0.222     & 0.156       \\
system 2 run16-Voting-1.1-SubtitleAndHfw-SentenceVec\_method\_2\_community     & 0.199        & 0.116          & 0.131         & 0.071           & 0.207     & 0.156       \\
system 2 run12-Jaccard-Focused-Voting-SentenceVec\_method\_2\_community        & 0.199        & 0.115          & 0.131         & 0.073           & 0.222     & 0.156       \\
system 2 run19-Voting-2.0-TextCNN-QD\_method\_1\_community                     & 0.198        & 0.114          & 0.135         & 0.072           & 0.226     & 0.156       \\
system 2 run23-Voting-2.0-Voting-QD\_method\_1\_community                      & 0.198        & 0.114          & 0.135         & 0.072           & 0.226     & 0.156       \\
system 2 run11-Jaccard-Focused-Voting-QD\_method\_1\_community                 & 0.197        & 0.108          & 0.134         & 0.069           & 0.220     & 0.154       \\
system 2 run7-Jaccard-Focused-SubtitleAndHfw-QD\_method\_1\_community          & 0.197        & 0.108          & 0.134         & 0.069           & 0.220     & 0.154       \\
system 1 Run 12                                                                & 0.184        & 0.111          & 0.192         & 0.110           & 0.194     & 0.151       \\
system 1 Run 2                                                                 & 0.183        & 0.111          & 0.192         & 0.112           & 0.193     & 0.150       \\
system 1 Run 6                                                                 & 0.183        & 0.111          & 0.192         & 0.112           & 0.193     & 0.150       \\
system 1 Run 14                                                                & 0.183        & 0.112          & 0.192         & 0.112           & 0.193     & 0.150       \\
system 2 run26-Word2vec-H-CNN-SubtitleAndHfw-SentenceVec\_method\_2\_community & 0.180        & 0.106          & 0.112         & 0.063           & 0.211     & 0.147       \\
system 1 Run 28                                                                & 0.167        & 0.104          & 0.192         & 0.108           & 0.194     & 0.150       \\
system 1 Run 22                                                                & 0.167        & 0.104          & 0.192         & 0.108           & 0.194     & 0.150       \\
system 1 Run 16                                                                & 0.167        & 0.104          & 0.192         & 0.108           & 0.194     & 0.150       \\
system 1 Run 24                                                                & 0.166        & 0.104          & 0.193         & 0.109           & 0.194     & 0.150       \\
system 2 run25-Word2vec-H-CNN-SubtitleAndHfw-QD\_method\_1\_community          & 0.151        & 0.097          & 0.126         & 0.069           & 0.201     & 0.138       \\
system 1 Run 13                                                                & 0.144        & 0.077          & 0.148         & 0.087           & 0.146     & 0.111       \\
system 1 Run 17                                                                & 0.119        & 0.063          & 0.149         & 0.095           & 0.128     & 0.098       \\
system 1 Run 11                                                                & 0.114        & 0.066          & 0.145         & 0.088           & 0.099     & 0.085       \\
system 1 Run 5                                                                 & 0.112        & 0.067          & 0.136         & 0.085           & 0.140     & 0.103       \\
system 1 Run 27                                                                & 0.110        & 0.064          & 0.136         & 0.089           & 0.142     & 0.098       \\
system 1 Run 25                                                                & 0.107        & 0.061          & 0.145         & 0.092           & 0.113     & 0.086       \\
system 1 Run 1                                                                 & 0.107        & 0.061          & 0.156         & 0.096           & 0.139     & 0.098       \\
system 1 Run 15                                                                & 0.105        & 0.062          & 0.128         & 0.080           & 0.097     & 0.077       \\
system 1 Run 9                                                                 & 0.101        & 0.063          & 0.147         & 0.091           & 0.121     & 0.091       \\
system 1 Run 29                                                                & 0.093        & 0.057          & 0.139         & 0.086           & 0.120     & 0.093       \\
system 1 Run 19                                                                & 0.091        & 0.056          & 0.157         & 0.083           & 0.113     & 0.085       \\
system 1 Run 23                                                                & 0.090        & 0.058          & 0.165         & 0.094           & 0.108     & 0.084       \\
system 1 Run 3                                                                 & 0.089        & 0.051          & 0.146         & 0.084           & 0.116     & 0.088       \\
system 1 Run 7                                                                 & 0.082        & 0.050          & 0.162         & 0.096           & 0.121     & 0.095       \\
system 1 Run 21                                                                & 0.075        & 0.050          & 0.109         & 0.063           & 0.121     & 0.083   \\
\hline   
\end{tabular}
}
\caption{
\scriptsize
Systems' performance for Task 2 ordered by their 
ROUGE--2(R--2) and ROUGE--SU4(R--SU4) $F_1$-scores. 
Each system's rank by their performance on the corresponding 
evaluation is shown in parentheses. Winning scores are bolded.}
\label{tab:task2}
\end{table}

\begin{figure}
  \centering
  \subfigure[]{\includegraphics[width=\textwidth]{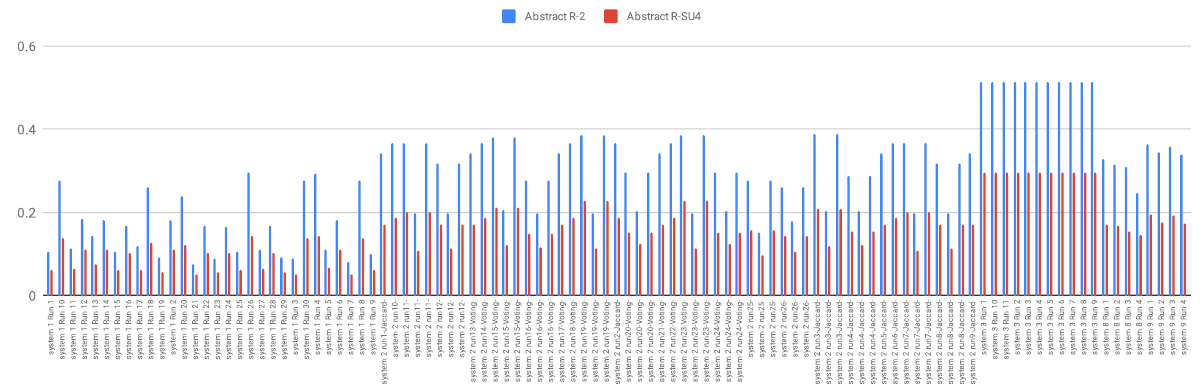}\label{fig:exph}}
  \subfigure[]{\includegraphics[width=\textwidth]{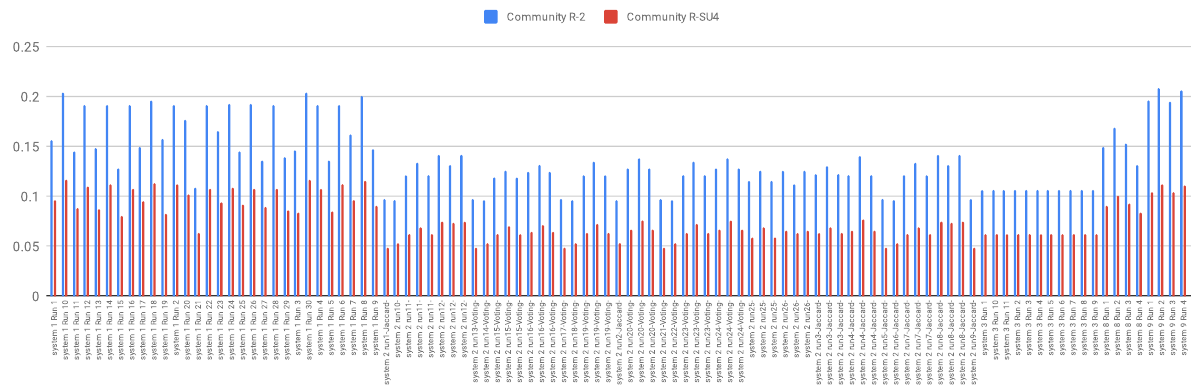}\label{fig:exgr}}
  \subfigure[]{\includegraphics[width=\textwidth]{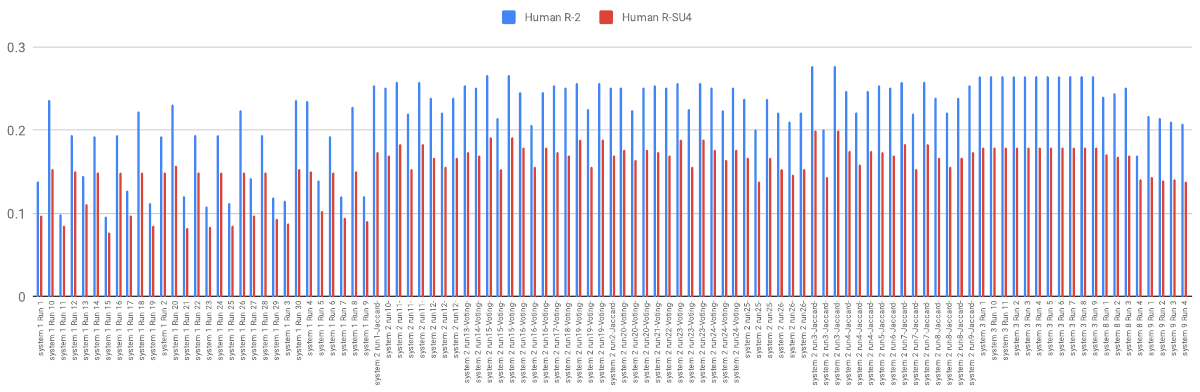}\label{fig:exgr}}
  \caption{Task 2 Performances on (a) Abstract, (b) Community and (c)
    Human summaries. Plots correspond to the numbers in Table~\ref{tab:task2}.}
  \label{fig:task2}
\end{figure}

For Task~1A, the best performance was shown by System~3 
(Team UoM)~\cite{system_3}. 
Their performance was closely followed by
System~12~\cite{system_12}. Both teams implemented 
deep learning-based systems.
One of the key goals of CL-SciSumm '19 was to boost 
performance of deep learning models by adding more 
training data. It is encouraging though not surprising to see 
the best performance from deep learning models. 
The third best system was system 2 (Team CIST-BUPT) which 
was also the best performer for Task~1B, the classification task. 
Second best performance Task~1B was by System~4 (Team IRIT-IRIS).\\ 

On the summarisation task, Task 2, System~3 (Team UoM) 
had the best 
performance against the abstract. 
System~2 (Team CIST-BUPT) had the best performance for
community and human summaries. Again, both are deep learning-based systems.
The additional 1000 summaries from SciSummnet as training 
data has resulted in the improved performance.
System~2 was the second against abstract summaries, and 
system~3 was the second against human summaries.

\section{Research questions and discussions}
For CL-SciSumm '19, we augmented the CL-SciSumm '18 training 
datasets for both Task 1a and Task 2 so that they have 
approximately 1000 data points as opposed to 40 in previous 
years. Specifically, for Task 1, we used the method proposed by \cite{nomoto2018resolving} to prepare noisy training data for 
about 1000 unannotated papers; for Task 2, we used the SciSummNet corpus proposed by \cite{yasunaga&al.aaai19.scisumm}.
For CL-SciSumm '19 we use the same blind test data used 
in CL-SciSumm '18.

Based on this we propose the following research questions to 
comparatively analyse results from CL-SciSumm '18 with 
those from CL-SciSumm '19.
The research questions we have are:\\
\textbf{RQ1.} Did data augmentation help systems 
achieve better performance? \\
The best Task~1a performance (sentence overlap $F_1$) 
this year is 0.126 from System~3~\cite{system_3} 
which is a deep learning system trained 
on augmented data. This is  
about 0.02 lower than the best CL-SciSumm'18 
system~\cite{wang2018nudt} which was at 0.145. 
It appears that the data augmentation has helped
deep learning methods. The only fully deep learning 
system from CL-SciSumm '18~\cite{de2018university}  
achieved 0.044. So, increasing training data is clearly  
the way forward. Traditional machine learning based 
systems such as~\cite{system_2} seem to  suffer from noise in the augmented data. 
We propose to use 
better data generation method that produces data cleaner 
than the naive similarity based cut-off 
method~\cite{nomoto2018resolving} used this time.

Note that there was no data augmentation to Task~1b. So, 
the performance of traditional methods across CL-SciSumm '18
and CL-SciSumm, '19 are largely the same.

The best on CL-SciSumm '19 Task 2 
performance on human written summaries 
on ROUGE-2 is 0.278 by \cite{system_2}.
This is higher than the best CL-SciSumm'18 
system which score 0.252~\cite{abura2018lastus}. 
This suggests that the additional 1000 ScisummNet 
summaries is useful to further performance.
It also indicates that SciSummNet relatively 
cleaner than the auto annotated data used for 
Task~1a.

\textbf{RQ2.} CL-SciSumm~'19 encouraged participants to 
use deep learning based methods; do they perform better 
than traditional machine learning methods?\\
In Task~1a the best performing CL-SciSumm~'19 system 
The best performing CL-SciSumm~'18 system~\cite{wang2018nudt} 
used  traditional models including random forests and 
ranking models trained on the CL-SciSumm '18 training data. 
This implies that for Task~1a, traditional 
models trained on clean data perform better than deep 
learning models trained on noisy data. However, if we 
look at CL-SciSumm '19 systems' performances, we notice 
that deep learning models perform better than 
traditional machine learning models when trained on the 
augmented data.

On Task~1b, systems using traditional methods perform  
better than deep learning systems. Note that the winner 
for Task~1a, System~3, is not the best system for Task~1b 
although they are not far behind. We also did not 
add any additional training data to Task~1b. So, we 
cannot rule out that deep learning systems will not 
perform better than traditional methods when trained 
on enough data.

On Task~2, the best performing system on human summaries, 
System~2, using neural representations trained on the 
1000 plus summaries, does the best with a ROUGE-2 score of 
0.278.  This is higher than CL-SciSumm~'18 top system using 
traditional methods. System~3, the second best Cl-SciSumm '19 
system an end-end deep learning model, with a score of 0.265  
is also higher than CLSciSumm '18 top system. With a score 
of  0.514 System~3 also improves the state-of-the-art agasint  
abstracts by 0.2 on ROUGE-2 score. System~3 is also the top 
system on community summaries with a ROUGE-2 score of 0.204.

In summary, deep learning models do well across the board for 
summaries. Traditional methods do better on Task~1a on 
small but clean training data. Deep learning methods take over
on large bu tnoisy data.




\section{Conclusion}
\label{s:conc}

Nine systems participated in CL-SciSumm 2019 shared tasks. The systems 
were provided with larger but noisy corpus with automatic annotation. Nearly
all the teams had neural methods and many employed transfer learning. 
Participants also experimented with the use of word embeddings trained 
on the shared task corpus, as well as on other domain corpora. We found 
that data augmentation for Task~1a may have helped deep learning models but 
not traditional machine learning methods. It also appears that deep learning 
methods perform better than traditional methods across the board when they 
have enough training data. We will explore methods to obtain cleaner 
training data for Task~1 without or with minimal human annotation effort.

We recommend that future approaches should go
beyond off-the-shelf deep learning methods, and also exploit the
structural and semantic characteristics that are unique to scientific
documents; perhaps as an enrichment device for word embeddings. 
The committee also observes that CL-SciSumm series over the past 5 years 
has catalysed research in the area of scientific document summarisation. 
We observe that a number of papers outside of the BIRNDL workshop  
published at prominent NLP and IR venues evaluate on the CL-SciSumm 
gold standard data. To create a reference corpus for the task was 
a key goal of the series. We have achieved this goal now. We will 
consider newer tasks to push the effort towards automated literature 
reviews. We will also consider switching the format of the shared evaluation 
from a shared task to a leaderboard to which systems can submit evaluations 
asynchronously throughout the year.

\renewcommand{\abstractname}{\ackname}
\begin{abstract}

We would like to thank SRI International  for their generous 
funding of CL-SciSumm '19 and BIRNDL '19. We thank 
Chan-Zuckerberg Initiative for sponsoring the invited talk.
We would also like to thank Vasudeva Varma and colleagues at
IIIT-Hyderabad, India and University of Hyderabad for their efforts in
convening and organizing our annotation workshops in 2016-17. We acknowledge 
the continued advice of Hoa Dang, Lucy Vanderwende and Anita de Waard 
from the pilot stage of this task. We would also like to thank Rahul Jha
and Dragomir Radev for sharing their software to prepare the XML
versions of papers. We are grateful to Kevin B. Cohen and colleagues
for their support, and for sharing their annotation schema, export
scripts and the Knowtator package implementation on the Protege
software -- all of which have been indispensable for this shared task.
\end{abstract}


%
\bibliographystyle{splncs03}
\bibliography{splncs}

\begin{thebibliography}{10}
\providecommand{\url}[1]{\texttt{#1}}
\providecommand{\urlprefix}{URL }

\bibitem{abura2018lastus}
Abura’ed, A., Bravo, A., Chiruzzo, L., Saggion, H.: Lastus/taln+ inco@
  cl-scisumm 2018-using regression and convolutions for cross-document semantic
  linking and summarization of scholarly literature. In: Proceedings of the 3nd
  Joint Workshop on Bibliometric-enhanced Information Retrieval and Natural
  Language Processing for Digital Libraries (BIRNDL2018). Ann Arbor, Michigan
  (July 2018) (2018)

\bibitem{system_8}
AbuRa’ed, A., Chiruzzo, L., Bravo, A., Saggion, H.: {LaSTUS-TALN+INCO @
  CL-SciSumm 2019}. In: BIRNDL2019 (2019)

\bibitem{de2018university}
De~Moraes, L.F., Das, A., Karimi, S., Verma, R.M.: University of houston@
  cl-scisumm 2018. In: BIRNDL@ SIGIR. pp. 142--149 (2018)

\bibitem{system_17}
Fergadis, A., Pappas, D., Papageorgiou, H.: {Siamese recurrent bi-directional
  neural network for scientific summarization @ CL-SciSumm 2019 }. In:
  BIRNDL2019 (2019)

\bibitem{jaidka2017overview}
Jaidka, K., Chandrasekaran, M.K., Jain, D., Kan, M.Y.: The cl-scisumm shared
  task 2017: Results and key insights. In: BIRNDL@ SIGIR (2). vol. 2002, pp.
  1--15. CEUR (2017)

\bibitem{jaidka2017insights}
Jaidka, K., Chandrasekaran, M.K., Rustagi, S., Kan, M.Y.: Insights from
  cl-scisumm 2016: the faceted scientific document summarization shared task.
  International Journal on Digital Libraries pp. 1--9 (2017)

\bibitem{jaidka2018overview}
Jaidka, K., Yasunaga, M., Chandrasekaran, M.K., Radev, D., Kan, M.Y.: The
  cl-scisumm shared task 2018: Results and key insights. In: BIRNDL@ SIGIR (2).
  vol. 2132, pp. 74--83. CEUR (2018)

\bibitem{sparckjones2007automatic}
Jones, K.S.: {Automatic summarising: The state of the art}. Information
  Processing and Management  43(6),  1449--1481 (2007)

\bibitem{system_12}
Kim, H., Ou, S.: {Ranking-based Identification of Cited Text with Deep Learning
  }. In: BIRNDL2019 (2019)

\bibitem{system_2}
Li, L., Zhu, Y., Xie, Y., Huang, Z., Liu, W., Li, X., Liu, Y.:
  {CIST@CLSciSumm-19: Automatic Scientific Paper Summarization with Citances
  and Facets}. In: BIRNDL2019 (2019)

\bibitem{lin2004rouge}
Lin, C.Y.: Rouge: A package for automatic evaluation of summaries. {Text
  summarization branches out: Proceedings of the ACL-04 workshop}  8 (2004)

\bibitem{liu2008correlation}
Liu, F., Liu, Y.: Correlation between rouge and human evaluation of extractive
  meeting summaries. In: Proceedings of the 46th Annual Meeting of the
  Association for Computational Linguistics on Human Language Technologies:
  Short Papers. pp. 201--204. Association for Computational Linguistics (2008)

\bibitem{system_1}
Ma, S., Zhang, H., Xu, T., Xu, J., Hu, S., Zhang, C.: {IR\&TM-NJUST @
  CLSciSumm-19}. In: BIRNDL2019 (2019)

\bibitem{MayrCJ17}
Mayr, P., Chandrasekaran, M.K., Jaidka, K.: Editorial for the 2nd joint
  workshop on bibliometric-enhanced information retrieval and natural language
  processing for digital libraries {(BIRNDL)} at {SIGIR} 2017. In: Proceedings
  of the 2nd Joint Workshop on Bibliometric-enhanced Information Retrieval and
  Natural Language Processing for Digital Libraries {(BIRNDL} 2017) co-located
  with the 40th International {ACM} {SIGIR} Conference on Research and
  Development in Information Retrieval {(SIGIR} 2017), Tokyo, Japan, August 11,
  2017. pp. 1--6 (2017), \url{http://ceur-ws.org/Vol-1888/editorial.pdf}

\bibitem{mayr16:_editorial}
Mayr, P., Frommholz, I., Cabanac, G., Wolfram, D.: {Editorial for the Joint
  Workshop on Bibliometric-enhanced Information Retrieval and Natural Language
  Processing for Digital Libraries (BIRNDL) at JCDL 2016}. In: Proc.\ of the
  Joint Workshop on Bibliometric-enhanced Information Retrieval and Natural
  Language Processing for Digital Libraries {(BIRNDL2016)}. pp. 1--5. Newark,
  NJ, USA (June 2016)

\bibitem{nakov2004citances}
Nakov, P.I., Schwartz, A.S., Hearst, M.: {Citances: Citation sentences for
  semantic analysis of bioscience text}. In: {Proceedings of the SIGIR'04
  workshop on Search and Discovery in Bioinformatics}. pp. 81--88 (2004)

\bibitem{nomoto2018resolving}
Nomoto, T.: Resolving citation links with neural networks. Frontiers in
  Research Metrics and Analytics  3, ~31 (2018)

\bibitem{system_4}
Pitarch, Y., Pinel-Sauvagnat, K., Hubert, G., Cabanac, G., ́elie
  Fraisier-Vannier, O.: {IRIT-IRIS at CL-SciSumm 2019: Matching Citances with
  their Intended Reference Text Spans from the Scientific Literature}. In:
  BIRNDL2019 (2019)

\bibitem{qazvinian2008scientific}
Qazvinian, V., Radev, D.: Scientific paper summarization using citation summary
  networks. In: {Proceedings of the 22nd International Conference on
  Computational Linguistics-Volume 1}. pp. 689--696. ACL (2008)

\bibitem{system_9}
Quatra, M.L., Cagliero, L., Baralis, E.: {Poli2Sum@CL-SciSumm 2019: identify,
  classify, and summarize cited text spans by means of ensembles of supervised
  models }. In: BIRNDL2019 (2019)

\bibitem{system_7}
Syed, B., Indurthi, V., Srinivasan, B.V., Varma, V.: {Transfer learning for
  effective scientific research comprehension}. In: BIRNDL2019 (2019)

\bibitem{wang2018nudt}
Wang, P., Li, S., Wang, T., Zhou, H., Tang, J.: Nudt@ clscisumm-18. In: BIRNDL@
  SIGIR. pp. 102--113 (2018)

\bibitem{yasunaga&al.aaai19.scisumm}
Yasunaga, M., Kasai, J., Zhang, R., Fabbri, A., Li, I., Friedman, D., Radev,
  D.: {ScisummNet}: A large annotated corpus and content-impact models for
  scientific paper summarization with citation networks. In: Proceedings of
  AAAI 2019 (2019)

\bibitem{system_3}
Zerva, C., Nghiem, M.Q., Nguyen, N.T., Ananiadou, S.: {UoM@CL-SciSumm 2019}.
  In: BIRNDL2019 (2019)

\end{thebibliography}
\end{document}